\documentclass{article} % For LaTeX2e
\PassOptionsToPackage{table,svgnames}{xcolor}
\usepackage{iclr2027_conference,times}

\usepackage{amsmath,amsfonts,bm}

\def\Figref#1{Figure~\ref{#1}}

\def\Secref#1{Section~\ref{#1}}
\def\eqref#1{equation~\ref{#1}}
\def\1{\bm{1}}

\DeclareMathAlphabet{\mathsfit}{\encodingdefault}{\sfdefault}{m}{sl}
\SetMathAlphabet{\mathsfit}{bold}{\encodingdefault}{\sfdefault}{bx}{n}

\usepackage{hyperref}
\usepackage{url}
\usepackage{booktabs} 
\usepackage{graphicx} 
\usepackage[table]{xcolor}
\usepackage{hyperref}
\usepackage{url}
\usepackage{graphicx}
\usepackage{booktabs}
\usepackage{amsmath}
\usepackage{amssymb}
\usepackage{multirow}
\usepackage{todonotes}
\usepackage{array}
\usepackage{xcolor}
\usepackage{colortbl}
\usepackage{pifont}

\usepackage{hyperref}
\usepackage{url}
\usepackage{booktabs}
\usepackage{wrapfig}
\usepackage{svg}
\usepackage{xcolor}
\usepackage{amsmath}
\usepackage{calligra}
\usepackage{listings}
\usepackage{colortbl}
\usepackage[most]{tcolorbox} % loads xcolor
\tcbuselibrary{listings,skins,breakable}
\usepackage[svgnames]{xcolor}
\usepackage[most]{tcolorbox}
\usepackage{fontawesome5} 

\usepackage{hyperref}
\usepackage{url}
\usepackage{graphicx}
\usepackage{booktabs}
\usepackage{amsmath}
\usepackage{amssymb}
\usepackage{multirow}
\usepackage{array}
\usepackage{xcolor}
\usepackage{colortbl}
\usepackage{pifont} % \ding{51} check, \ding{55} cross
\usepackage{soul} 

\usepackage{float}

\author{%
Raja Kumar \textsuperscript{1}\thanks{work done during internship at Amazon AGI.\newline
\begingroup
\setbox0=\hbox{$^{*}$}%
\hspace*{\dimexpr1.8em-\wd0\relax}%
Correspondence: \href{mailto:kumarraj@usc.edu}{\texttt{kumarraj@usc.edu}} and \href{mailto:rajatkon@amazon.com}{\texttt{rajatkon@amazon.com}}.\endgroup},
Rajat Koner \textsuperscript{2},
Ritwick Chaudhry \textsuperscript{2}, 
Zhuowei Li \textsuperscript{2}, %\quad
Nishant Sankaran \textsuperscript{2}, \\
\textbf{Yifan Xing\textsuperscript{2}} \\[0.5em]
\normalfont\textsuperscript{1}University of Southern California, 
\normalfont\textsuperscript{2}Amazon AGI
}

\hypersetup{%
  pdfauthor={Raja Kumar, Rajat Koner, Ritwick Chaudhry, Zhuowei Li, Nishant Sankaran, Yifan Xing}
}

\definecolor{lightgray}{gray}{0.92}

\definecolor{posgreen}{RGB}{0,140,0}
\definecolor{negred}{RGB}{200,0,0}
\newcommand{\dpos}[1]{\textcolor{posgreen}{#1}}   % positive deltas
\newcommand{\dneg}[1]{\textcolor{negred}{#1}} 
\newcommand{\mtdname}{{SPLIT-RL}}
\newcommand{\claname}{{CLA-GRPO}}

\title{SPLIT-RL: Staged Perception–Language Reasoning Training with Claim-Level Advantages}

\iclrpreprintcopy 
\begin{document}

\maketitle

\begin{abstract}
Vision--Language (VL) reasoning requires a model to both extract relevant and accurate
information from an image (\emph{visual reasoning}, VR), and to infer the answer
from it (\emph{language reasoning}, LR). Reinforcement learning with verifiable
rewards typically trains both through a single chain-of-thought
with a final-answer reward. This gives every CoT token the same sequence-level
advantage, failing to distinguish capability specific errors.  
We propose %\textbf{S}taged \textbf{P}erception--\textbf{L}anguage reason\textbf{I}ng
%\textbf{T}raining with \textbf{R}einforcement \textbf{L}earning
\textbf{SPLIT-RL}, 
a staged post-training approach that trains VR and LR in disjoint phases. 
Because a group's rollouts differ along one capability at a time, the
group-relative advantage isolates it, and each phase is optimized using phase-specific reward.
% We propose \textbf{S}taged \textbf{P}erception--\textbf{L}anguage reason\textbf{I}ng
% \textbf{T}raining with \textbf{R}einforcement \textbf{L}earning
% (\textbf{SPLIT-RL}), a staged post-training approach
% that performs disjoint capability-specific (VR/LR) training, making group-relative rollouts differ in specific capability and also allowing capability-specific rewards. 
%separates the two capabilities within one shared policy. In a VR phase, the model describes the visual content relevant to the question; in an LR phase, the same model answers from this description alone, without the image. The description thus becomes an information bottleneck, so final-answer correctness verifies perception without any ground-truth captions, and the two phases can be trained one after the other, each against its own reward. 
We further introduce
Claim-Level Advantage (\claname{}), which decomposes VR-phase rollouts into atomic visual claims and provides a fine-grained advantage at claim level based on visual-type group formation. Although trained in two phases, trained policy is evaluated like GRPO model, with a single CoT call at inference time. Under this protocol, \mtdname{} improves average accuracy over GRPO by 1.4--6.1 points across Qwen3-VL
models from 2B to 30B-A3B and InternVL3.5-8B.
Evaluating each capability using an oracle based diagnostic shows that answer-only GRPO leaves perception unchanged, whereas \mtdname{} improves both VR and LR.

%how well the model describes an image and how well it reasons from a description.
\end{abstract}

% \input{1_introduction}

%=====================================================================
\section{Introduction}
%=====================================================================

\begin{figure}[t]
\centering
\includegraphics[width=\textwidth]{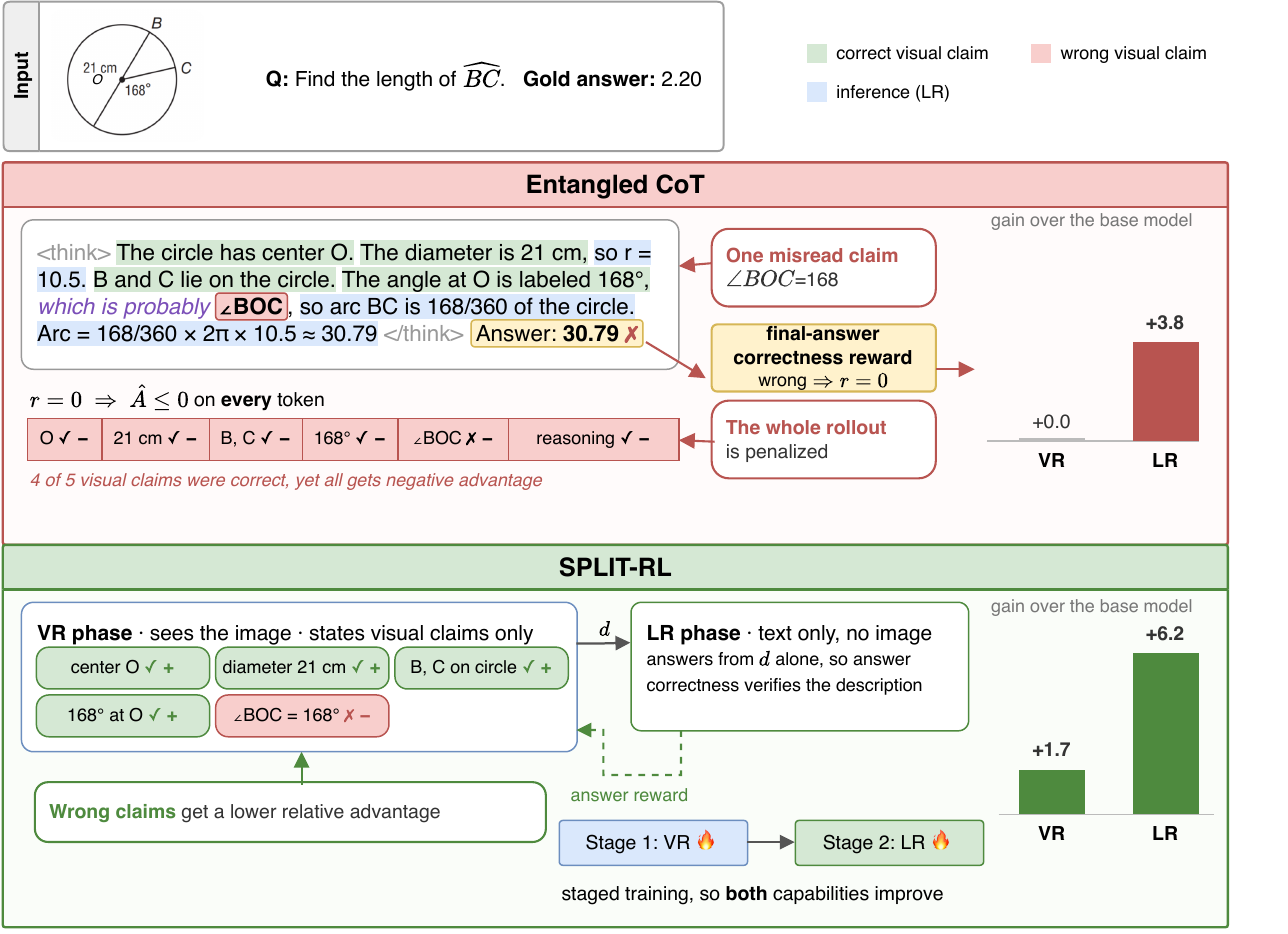}

\caption{\textbf{Entangled CoT vs.\ SPLIT-RL.}
\textbf{Top:} One misread visual claim produces a wrong answer, assigning every token the same nonpositive advantage including four correct claims. With a fixed oracle (Section~4.3.2), answer-only training improves LR (+3.8) but not VR (+0.0).
\textbf{Bottom:} SPLIT-RL separates visual description from text-only reasoning, using answer correctness to assess description sufficiency. CLA-GRPO assigns claim-level advantages, and training VR before LR improves both capabilities (+1.7 VR, +6.2 LR).}

\label{fig:teaser}
\end{figure}

% --- P1: VL reasoning has an inherent two-stage structure (VR -> LR) ---
Vision--Language (VL) reasoning pairs an image with a question that cannot be answered from text knowledge alone. Solving it involves two capabilities: interpreting the question and extracting the visual information it requires - Visual Reasoning (VR), followed by inferring the answer from that information - Language Reasoning (LR). These capabilities underpin reliable multimodal systems such as chart and diagram QA \citep{masry2022chartqa}, robotic manipulation \citep{brohan2023rt} and medical imaging \citep{li2025vision}. Modern Vision Language Models (VLMs) \citep{qwen3vl2025} answer such questions by generating a Chain-of-Thought (CoT)~\citep{wei2022cot} followed by a final answer, and are commonly post-trained using reinforcement learning with verifiable rewards (RLVR) \citep{lambert2024tulu}. %, using a final-answer reward optimized by GRPO ~\citep{shao2024deepseekmath}.

% --- P2: entangled CoT + what recent work focuses on ---
In this setup, a single CoT stream interleaves VR and LR, which we call the \emph{entangled CoT} and a final-answer correctness reward is optimized using GRPO \citep{deepseekr1}. GRPO improves performance, but the entangled CoT makes credit assignment coarse at two levels. First, because each rollout mixes VR and LR, the group-relative advantage is not capability-specific: it cannot directly optimize either capability in isolation. Second, a rollout may contain multiple visual claims and reasoning steps, some correct and some not, yet GRPO assigns a single sequence-level advantage based on the final-answer reward. As a result, incorrect claims may be reinforced when the final answer is correct, while correct claims may be penalized whenever answer is wrong (Figure~\ref{fig:teaser}, top). The entanglement also makes capability-specific rewards hard to define in the first place: with VR and LR interleaved token by token, there is no clean segment on which to place a perception reward.

Recent works \citep{zhang2026perceptual, xiao2026perception} add perception-specific rewards and training strategies to address this limitation. PEARL~\citep{zhang2026perceptual} builds a perception checklist for verification; Perception-R1~\citep{xiao2026perception} prompts an LLM to generate perception claims and verifies the CoT against them, and Vision-SR1~\citep{li2026visionsr1} re-prompts the policy on its own perception, without the image, to obtain a self-reward. However, these methods either require an offline data-curation stage or optimize perception jointly with the final answer in a single response, and all retain sequence-level perception rewards. 

% --- P4: what we propose -> why two stages -> name ---
To this end, we propose \textbf{S}taged \textbf{P}erception--\textbf{L}anguage reason\textbf{I}ng \textbf{T}raining with \textbf{R}einforcement \textbf{L}earning (\textbf{\mtdname{}}), which sequentially trains VR and LR as two disjoint stages of a single policy (\Figref{fig:teaser}, bottom). In the VR stage, the model sees the image and the query and performs VR to generate query-relevant visual description, with no reasoning and no answer. In the LR stage, the same model receives only this description and the query and reasons to the answer. %We hypothesize that this separation strengthens each capability in its own way.
Because the reasoner never sees the image, the description becomes an information bottleneck: a rollout succeeds only if the description alone suffices to answer, which pushes the model toward more complete and accurate perception without ground-truth descriptions or VR-specific data. Conversely, removing the image forces the reasoner to derive the answer from stated evidence alone, grounding every step in text rather than in shortcuts from the image. 

% --- P5: VR rewards -- completeness (recall) + correctness (CLA-GRPO), high level ---
While staged training enables disjoint VR/LR optimization and capability-specific relative advantage computation, credit assignment within the VR description remains coarse. \mtdname{} therefore introduces Claim-Level Advantage (\claname{}), which brings fine-grained credit assignment to the VR stage. \claname{} decomposes the description into atomic visual claims, verifies each against the image, and groups claims of the same type (e.g., color or size) into sibling groups, so that each claim's advantage is computed relative to claims of the same kind. This provides fine-grained supervision for the correctness of individual visual claims. However, correctness alone does not ensure that the description contains all visual information required to answer the query. We therefore complement VR stage with a completeness reward named recall reward that encourages the VR description to capture all query-relevant evidence needed by the LR stage.

% --- P6: results ---
On six visual benchmarks, evaluated under the standard GRPO setting of single stage CoT inference, \mtdname{} outperforms entangled CoT GRPO across five backbones, by $+1.4$ to $+6.1$ average points, including $+2.7$ on Qwen3-VL-8B. Using our diagnostic probe to measure VR and LR improvement, we show that \mtdname{} improves LR by $+6.2$ over the base model, $+2.4$ over GRPO, and VR by $+1.7$, whereas GRPO leaves VR unchanged (see figure \ref{fig:teaser}). 

In summary, our main contributions are:
\begin{itemize}
  \setlength{\parskip}{0pt}
  \item We propose \mtdname{}, a staged post-training approach that disentangles VR and LR within a single policy and trains them in sequence, using the description as a bottleneck so that answer correctness verifies perception.
  \item We make VR training capability-specific by rewarding description completeness through a recall reward and correctness through claim-level advantages from \claname{}.
  \item We show consistent gains over GRPO across five backbones spanning model scales, architectures, and families, with extensive ablations analyzing capability gains.
\end{itemize}

%=====================================================================
\section{Related Work}
\label{sec:related}
%=====================================================================

%---------------------------------------------------------------------
\paragraph{Post-training RL for vision--language reasoning}
%---------------------------------------------------------------------
RLVR \citep{lambert2024tulu} has become the standard recipe for eliciting CoT reasoning from language models. DeepSeek-R1 ~\citep{deepseekr1} further made the training efficient by using GRPO. A rapid line of work has
since transferred this recipe to VLMs. Initial works such as Vision-R1~\citep{huang2025visionr1} and
R1-OneVision~\citep{yang2025r1onevision} pair a multimodal chain-of-thought
cold start with GRPO, while MM-Eureka~\citep{meng2025mmeureka} and
LMM-R1~\citep{peng2025lmmr1} show that rule-based RL alone yields stable gains on
multimodal math.
OpenVLThinker~\citep{deng2025openvlthinker} alternates SFT and RL, and VL-Rethinker~\citep{wang2025vlrethinker} and R1-VL~\citep{zhang2025r1vl} target reflection and step-wise rewards. 
RLVR has also been specialized to perception
tasks such as fine-grained reasoning, grounding, detection and
segmentation~\citep{kumar2026dive,shen2025vlmr1,liu2025segzero}, to
video~\citep{feng2025videor1}, medical imaging~\citep{lai2025medr1} and GUI
agents~\citep{luo2025guir1,lu2025uir1}. Across this breadth the interface is unchanged: a single entangled CoT interleaves perception and reasoning, trained by final-answer correctness alone. This is the setting we revisit in this work. 
%---------------------------------------------------------------------
\paragraph{Capability specific VLM post-training}
\label{sec:related-perception}
%---------------------------------------------------------------------

A growing body of work ~\citep{wu2026seeing,wang2025papo} has shown that VL reasoning interleaves VR and LR capabilities and final answer reward is insufficient. Process reward models \citep{lightman2023verify} score each intermediate step instead of only the outcome and VisualPRM~\citep{wang2025visualprm}, MM-PRM~\citep{du2025mmprm} and VRPRM~\citep{chen2025vrprm} train multimodal PRMs for vision--language reasoning. However, these methods require step-level annotations to train and are  most often inferred for best-of-$N$ selection at inference. Another line of work include curriculum learning where model is trained from from easy to hard data~\citep{deng2025curriculumrl}
or by 
perception to reasoning data \citet{wu2026seeing}. These stage data and therefore depend on curated corpora. We instead stage capability optimization, with a verifiable reward attached to each stage. Another family uses tools to revisit the image while reasoning such as sketching on it~\citep{hu2024sketchpad}, zooming and cropping~\citep{zheng2025deepeyes,zhang2025chainoffocus}, running pixel-space operations~\citep{wang2025pixelreasoner}, or invoking external visual tools~\citep{su2025openthinkimg}. These methods improve access to visual evidence, but come at the cost of additional compute \citep{wang2026moca}. We keep a pure-text interface and instead improve inherent capability. 
Our component-level signal is related in spirit to atomic fact verification for text generation~\citep{min2023factscore}. 
Similar to our work, MoCA~\citep{wang2026moca} and Vision-SR1 \citep{li2026visionsr1} use perception as a bottleneck to improve VR capability of the model. MoCA uses an external LLM while Vision-SR1 uses the same model as verifier. However, they still perform joint optimization and their VR specific reward remains coarse at sequence level unlike \mtdname{} which performs a staged optimization and provides a fine-grained advantage at claim level using \claname{}.

%=====================================================================
\section{\mtdname{}}
\label{sec:method}
%=====================================================================

\begin{figure}[t]
\centering
\includegraphics[width=\textwidth]{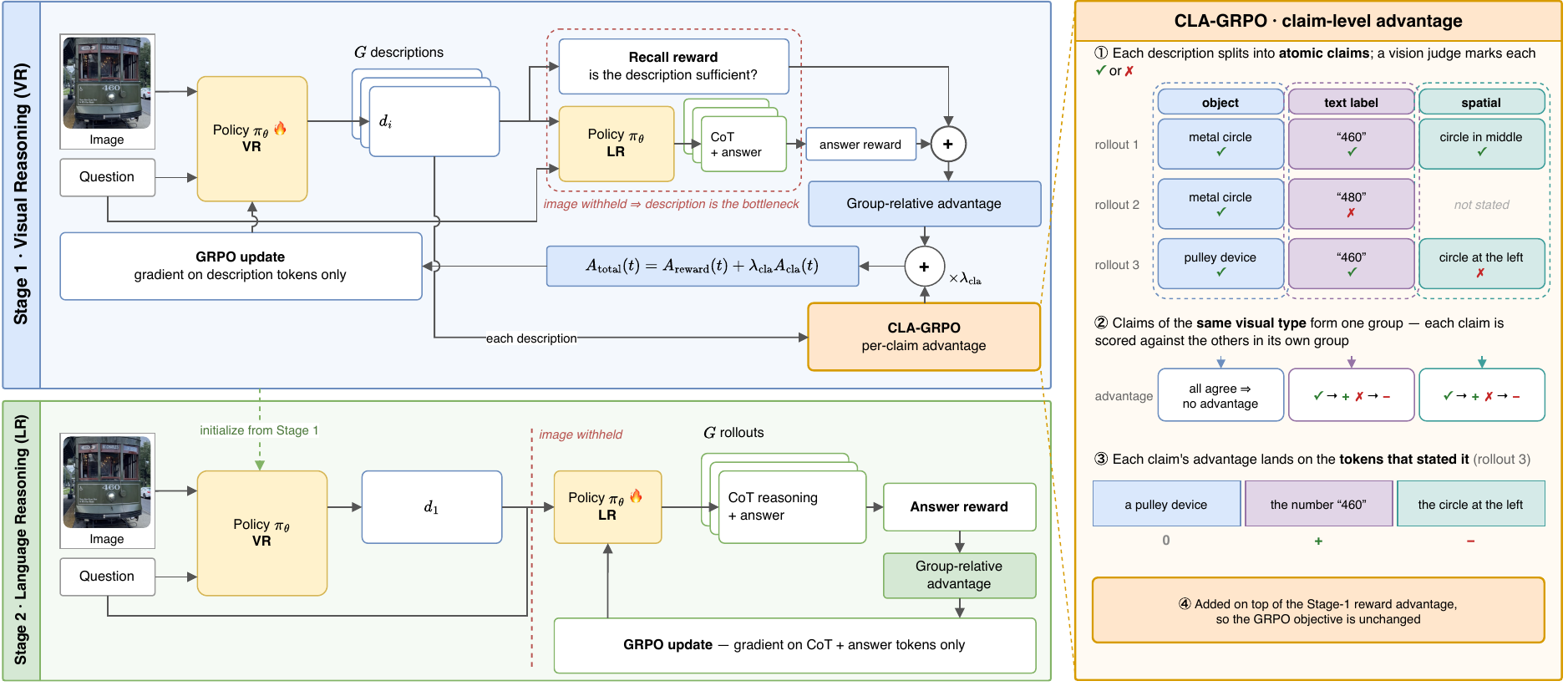}

\caption{\textbf{SPLIT-RL.} One policy is trained in two stages.
\textbf{Stage 1 (VR):} Samples $G$ descriptions per image--question pair and updates description tokens using recall, answer-correctness, and claim-level signals.
\textbf{Stage 2 (LR):} Starting from Stage 1, generates one description $d$ and samples $G$ reasoning--answer rollouts from $(Q,d)$ without the image; GRPO updates only reasoning and answer tokens.
\textbf{Right:} CLA-GRPO verifies atomic claims against the image, groups them by visual type across rollouts, and assigns claim-level advantages to their tokens.}

\label{fig:pipeline}
\end{figure}

We present \mtdname{}, a staged post-training approach that trains VR and LR within a single policy (\Figref{fig:pipeline}). We first discuss the disjoint VR and LR formulation(\Secref{sec:disjoint}). Then we present our staged training approach (\Secref{sec:staged}). Finally, we introduce VR- specific signals, CLA-GRPO and the recall reward, in Section (\ref{sec:vr-signals})  \claname{} and recall reward.

%---------------------------------------------------------------------
% \subsection{Preliminaries}
% \label{sec:prelim}
%---------------------------------------------------------------------

%---------------------------------------------------------------------
\subsection{Disjoint visual and language reasoning}
\label{sec:disjoint}
%---------------------------------------------------------------------
Let $I$ be an image, $Q$ a question and $a^\star$ the gold answer and $\pi_{\theta}$ is the policy itself. Standard GRPO~\citep{shao2024deepseekmath}
samples a group of $G$ rollouts from the policy,
$\{o_1,\dots,o_G\}\sim\pi_{\theta_{\text{old}}}(\cdot\mid I,Q)$, scores each one
by final-answer correctness $r_i = r(a^\star, o_i)\in\{0,1\}$, and computes the
advantage from the group statistics,
\begin{equation}
  \hat{A}_i \;=\; \frac{r_i - \operatorname{mean}(\{r_1,\dots,r_G\})}
                       {\operatorname{std}(\{r_1,\dots,r_G\})},
  \label{eq:grpo-adv}
\end{equation}
which is then optimized with the clipped surrogate objective \ref{eq:grpo}. The same scalar $\hat A_i$ is broadcast uniformly over all tokens of $o_i$, so tokens that perceive and tokens that reason receive identical credit, and the reward cannot tell which capability failed.

In \mtdname{}, we split a single generation into two phases performed by the same policy $\pi_\theta$. %\footnote{Prompts for both phases are given in Appendix~\ref{app:prompts}.}. 
In the VR phase, $\pi_\theta$ acts as a \emph{describer}: given the image $I$ and question $Q$, it produces a description $d$ of the visual content relevant to $Q$, with no reasoning and answer. In the LR phase, the same $\pi_\theta$ acts as a \emph{reasoner}: it receives only $Q$ and $d$ (no image) and derives the answer $\hat a$, leading to \eqref{eq:factorization} 
\begin{equation}
  d \sim \pi_\theta(\cdot \mid I, Q), \qquad
  \hat a \sim \pi_\theta(\cdot \mid Q, d), \qquad
  r_{\text{ans}}(d) = r(a^\star, \hat a) \in \{0,1\}.
  \label{eq:factorization}
\end{equation}

\paragraph{Description as a bottleneck.} 

Because the reasoner is conditioned on $d$ alone, any visual fact that $d$ omits---or states incorrectly---cannot be recovered downstream. The answer reward $r_{\text{ans}}(d)$ therefore measures whether $d$ is \emph{sufficient} for $Q$: a verifiable signal for perception that needs no caption-level supervision, turning every $(I, Q, a^\star)$ triple into VR training data. Crucially, the reasoner that supplies this signal \emph{is} $\pi_\theta$ itself. The verifier costs no extra parameters, and it grades each description against what the model can actually do with it, and the sufficiency bar is set by the model's own reasoning rather than an external judge.

%---------------------------------------------------------------------
\subsection{Staged training with capability-specific rollouts}
\label{sec:staged}
%---------------------------------------------------------------------

The factorization in \eqref{eq:factorization} lets us optimize VR and LR in sequence
rather than jointly. Both stages update the same parameters $\theta$ with the
objective \eqref{eq:grpo}; they differ in what is sampled, what is rewarded, and
which tokens receive the gradient.

\paragraph{Stage 1: VR.} For each $(I,Q)$, the policy samples $G$ descriptions
$(d_1,\dots,d_G) \sim \pi_\theta(\cdot \mid I, Q)$, and each is scored by the same policy reasoner, $\hat a_i \sim \pi_\theta(\cdot \mid Q, d_i)$, to obtain the bottleneck answer
reward $r_{\text{ans},i}$. The advantage is computed over $G$ sibling descriptions
and applied only to description tokens; the answer tokens act as part of the verifier and
receive no gradient. \Secref{sec:vr-signals} adds two further signals to this stage.

\paragraph{Stage 2: LR.} Stage~2 is initialized from the Stage-1 policy. The trained
describer produces one on-policy description $d \sim \pi_\theta(\cdot \mid I, Q)$,
and the policy samples $G$ answers conditioned on it,
$(\hat a_1,\dots,\hat a_G) \sim \pi_\theta(\cdot \mid Q, d)$, each rewarded by
$r(a^\star, \hat a_i)$. The advantage is applied only to reasoning and answer tokens.

Staging gives each capability a learning signal of its
own. First, rollouts differ only in the capability under training: Stage-1 rollouts
share the reasoner and differ only in what was described, while Stage-2 rollouts share
the description and differ only in how it was reasoned over, so the group-relative
baseline of \eqref{eq:grpo-adv} isolates one capability instead of averaging over
both. Second, sequencing lets each stage act on the other's improved behavior: Stage~2 trains the reasoner on the descriptions the tuned describer actually produces. 

%---------------------------------------------------------------------
\subsection{Completeness and correctness signals for the VR stage}
\label{sec:vr-signals}
%---------------------------------------------------------------------

The bottleneck reward is coarse and indirect: it is a single scalar per rollout, and as noted in \Secref{sec:disjoint}, it is set by the reasoner: a sufficient description can still draw a wrong answer, so a low reward caused by a reasoning failure may therefore be mistaken for a perception failure. We therefore add signals computed directly from the description. A description can fail in two ways: it can \emph{omit} a visual fact the question needs, or \emph{misstate} one. We address each with its own signal, a
recall reward for completeness (\Secref{sec:recall}) and a claim-level advantage for
correctness (\Secref{sec:cla}).

\subsubsection{Completeness: recall reward}
\label{sec:recall}

Completeness is a property of the description, not of the answer, so we score it with a judge that sees the description but never the image. A text-only judge $\mathcal{J}_{\text{recall}}$ is provided $(Q, d)$,
and reports whether $d$ alone states every visual fact needed to answer $Q$,
\begin{equation}
  r_{\text{recall}} = \mathcal{J}_{\text{recall}}(Q, d) \in \{0,1\}.
\end{equation}
Withholding the image is what makes this a completeness signal: the judge can call $d$
sufficient only if the description itself carries the evidence, so a description that
omits a needed fact cannot be rescued by the judge inspecting the image. The overall VR-stage reward is then
\begin{equation}
  r_{\text{overall}} = \lambda_{\text{recall}}\, r_{\text{recall}}
    + \lambda_{\text{ans}}\, r_{\text{ans}},
  \label{eq:combined-reward}
\end{equation}
with $r_{\text{ans}}$ as the bottleneck reward of \eqref{eq:factorization}. 

We keep both because they judge sufficiency differently. $r_{\text{recall}}$ is direct
and unconfounded by reasoning, but a \emph{proxy}: the judge never sees $a^\star$ and
scores $d$ against its own notion of what is needed. $r_{\text{ans}}$ is
\emph{grounded}---anchored to $a^\star$ and to the same $\pi_\theta$ that consumes $d$---so it optimizes the describer for the model's own reasoning rather
than the judge's. Its cost is the confound above: a right description with faulty
reasoning scores $0$. The two are therefore paired, not substituted.

The recall judge also keeps the VR phase in its role: a description that states or
derives the answer is scored $r_{\text{recall}}=0$ even when sufficient. This closes an
obvious shortcut---writing the answer into $d$ would otherwise make $r_{\text{ans}}$
trivially high without any real perception---and keeps the bottleneck intact.
In Appendix \Figref{fig:vrreward} shows an example.

\begin{table}[t]
\centering
\small
\caption{\textbf{Main results.} Accuracy (\%) on six benchmarks.
$\Delta$ rows show percentage-point gains over GRPO for the same backbone.
Bold marks the best accuracy per column within each backbone.}
\label{tab:main}
\newcommand{\up}[1]{{\scriptsize\dpos{$+#1$}}}
\newcommand{\dn}[1]{{\scriptsize\dneg{$-#1$}}}
\newcommand{\bb}[2]{%
  \cellcolor{white}\multirow{-#1}{*}{\shortstack[l]{#2}}}
\setlength{\tabcolsep}{3.5pt}
\resizebox{\textwidth}{!}{%
\begin{tabular}{llccccccc}
\toprule
Backbone & Method & MVista & MVision & MVerse & WeMath
  & LogicVista & DynaMath & \textbf{AVG} \\
\midrule
 & Base & 60.3 & 18.8 & 38.8 & 56.5 & 34.2 & 55.9 & 44.1 \\
 & GRPO & 60.6 & 35.9 & 44.4 & 54.1 & 36.7 & 56.5 & 48.0 \\
\rowcolor{lightgray}
 & \mtdname{}
 & \textbf{65.4} & \textbf{42.8} & \textbf{46.7}
 & \textbf{62.1} & \textbf{44.1} & \textbf{63.8}
 & \textbf{54.1} \\
\bb{4}{Qwen3-VL\\2B} & $\Delta$ vs.\ GRPO
 & \up{4.8} & \up{6.9} & \up{2.3}
 & \up{8.0} & \up{7.4} & \up{7.3} & \up{6.1} \\
\midrule
 & Base & 77.1 & 59.2 & 61.7 & 75.6 & \textbf{63.1} & 73.4 & 68.4 \\
 & GRPO & 78.0 & 61.8 & 66.4 & 79.8 & 56.4 & 74.7 & 69.5 \\
\rowcolor{lightgray}
 & \mtdname{}
 & \textbf{78.7} & \textbf{68.4} & \textbf{69.3}
 & \textbf{84.0} & 62.6 & \textbf{76.1}
 & \textbf{73.2} \\
\bb{4}{Qwen3-VL\\4B} & $\Delta$ vs.\ GRPO
 & \up{0.7} & \up{6.6} & \up{2.9}
 & \up{4.2} & \up{6.2} & \up{1.4} & \up{3.7} \\
\midrule
 & Base & 77.2 & 63.2 & 63.8 & 79.9 & 60.4 & 74.4 & 69.8 \\
 & GRPO & 77.9 & 64.5 & 67.8 & 84.6 & 65.6 & 75.3 & 72.6 \\
 & VLM-CapCurr~\citep{wu2026seeing}
 & \textbf{80.0} & 44.7 & 67.6 & 81.2 & 60.9 & 74.6 & 68.2 \\
 & PEARL~\citep{zhang2026perceptual}
 & 77.9 & 58.2 & 68.9 & 84.8 & 61.5 & 72.8 & 70.7 \\
\rowcolor{lightgray}
 & \mtdname{}
 & 79.8 & \textbf{69.4} & \textbf{69.8}
 & \textbf{87.4} & \textbf{66.0} & \textbf{79.1}
 & \textbf{75.3} \\
\bb{6}{Qwen3-VL\\8B} & $\Delta$ vs.\ GRPO
 & \up{1.9} & \up{4.9} & \up{2.0}
 & \up{2.8} & \up{0.4} & \up{3.8} & \up{2.7} \\
\midrule
 & Base & 79.6 & 59.2 & 65.7 & 75.5 & 60.4 & 78.1 & 69.8 \\
 & GRPO & \textbf{80.8} & 64.9 & \textbf{74.7}
 & 76.8 & 64.1 & 75.1 & 72.7 \\
\rowcolor{lightgray}
 & \mtdname{}
 & 80.0 & \textbf{66.4} & 73.0
 & \textbf{80.3} & \textbf{64.2} & \textbf{80.9}
 & \textbf{74.1} \\
\bb{4}{Qwen3-VL\\30B-A3B} & $\Delta$ vs.\ GRPO
 & \dn{0.8} & \up{1.5} & \dn{1.7}
 & \up{3.5} & \up{0.1} & \up{5.8} & \up{1.4} \\
\midrule
 & Base & 79.2 & 55.9 & 60.9 & 83.9 & 57.3 & 70.5 & 68.0 \\
 & GRPO & 79.4 & 62.2 & 63.2 & 81.6 & 57.3 & 73.6 & 69.5 \\
\rowcolor{lightgray}
 & \mtdname{}
 & \textbf{80.6} & \textbf{63.5} & \textbf{64.0}
 & \textbf{84.6} & \textbf{60.1} & \textbf{77.1}
 & \textbf{71.6} \\
\bb{4}{InternVL3.5\\8B} & $\Delta$ vs.\ GRPO
 & \up{1.2} & \up{1.3} & \up{0.8}
 & \up{3.0} & \up{2.8} & \up{3.5} & \up{2.1} \\
\bottomrule
\end{tabular}%
}
\end{table}

\subsubsection{Correctness: claim-level advantage}
\label{sec:cla}

Both rewards in \eqref{eq:combined-reward} are \emph{holistic}: each returns a single
scalar for the whole description, so GRPO assigns the rollout one advantage and credits
every claim's tokens identically (\Figref{fig:teaser}, top). A description, though, is
many claims, each right or wrong on its own. When one misstated claim sits among many
correct ones, a single shared advantage cannot push it down without pushing the correct
claims down with it. \claname{} instead verifies each claim and gives it its own
advantage, acting directly on the tokens that stated it.

\paragraph{Decomposition and verification.} Each sampled description $d_i$ is
segmented into an ordered sequence of atomic claims, with the character
span of each retained,
\begin{equation}
  d_i \longrightarrow (c_{i1}, c_{i2}, \dots, c_{in_i}),
  \qquad c_{ij} \text{ spanning characters } [\alpha_{ij}, \beta_{ij}).
\end{equation}
A vision judge $\mathcal{J}_{\text{cla}}$ labels every claim with a correctness, a
relevance and a visual-type verdict,
\begin{equation}
  \mathcal{J}_{\text{cla}}(I, Q, c_{ij}) = (y_{ij},\, \ell_{ij},\, \tau_{ij}),
  \qquad
  y_{ij} \in \{0,1\},\quad
  \ell_{ij} \in \{\textsc{low}, \textsc{high}\},
\end{equation}
where $y_{ij}$ says whether the claim is faithful to the image, $\ell_{ij}$ its
relevance for answering $Q$, and
$\tau_{ij} \in \{\text{object presence},\ \text{count},\
\text{color/material},\ \text{spatial position},\ \text{text label},\
\text{geometric relation},\ \dots\}$ names the kind of visual property being
claimed. The full judge prompt is given in Appendix~\ref{app:judge-prompts}.

\paragraph{Claim-level advantage.} A rollout now carries a \emph{set} of verdicts,
for which the group-relative estimator \eqref{eq:grpo-adv} is not defined. \claname{} treats each
visual type $\tau$ as one group, so each claim is scored against what the
sibling rollouts said about \emph{the same kind of visual property on the same
image}. Concretely, the \textsc{high}-relevance claims of type $\tau$ made by the
$G$ sibling rollouts of a prompt form one sibling group,
\begin{equation}
  \mathcal{G}^{\tau} = \{\, (i,j) \;:\; \tau_{ij} = \tau,\ \ \ell_{ij} = \textsc{high} \,\},
  \label{eq:cla-group}
\end{equation}
and each claim's own correctness $y_{ij}$ is its reward. Applying the
group-relative estimator of \eqref{eq:grpo-adv} \emph{within a type} gives a
claim-level advantage
\begin{equation}
  \hat{A}_{ij} =
  \begin{cases}
    \dfrac{y_{ij} - \mu^{\tau_{ij}}}{\sigma^{\tau_{ij}}}, &
      (i,j) \in \mathcal{G}^{\tau_{ij}} \ \text{and}\ \sigma^{\tau_{ij}} > 0,\\[0.9em]
    0, & \text{otherwise},
  \end{cases}
  \qquad
  % \mu^{\tau} = \operatorname{mean}_{(i,j) \in \mathcal{G}^{\tau}} y_{ij},
  % \quad
  % \sigma^{\tau} = \operatorname{std}_{(i,j) \in \mathcal{G}^{\tau}} y_{ij}.
  \begin{aligned}
    \mu^{\tau}    &= \operatorname{mean}_{(i,j) \in \mathcal{G}^{\tau}} y_{ij},\\[0.2em]
    \sigma^{\tau} &= \operatorname{std}_{(i,j) \in \mathcal{G}^{\tau}} y_{ij}.
  \end{aligned}
  \label{eq:cla-adv}
\end{equation}

Credit is thus assigned per claim, not per rollout. Low-relevance claims,
and types with a single claim or no disagreement among the siblings, receive zero
advantage rather than an undefined normalization.

\begin{figure}[t]
\centering
\includegraphics[width=0.9\linewidth]{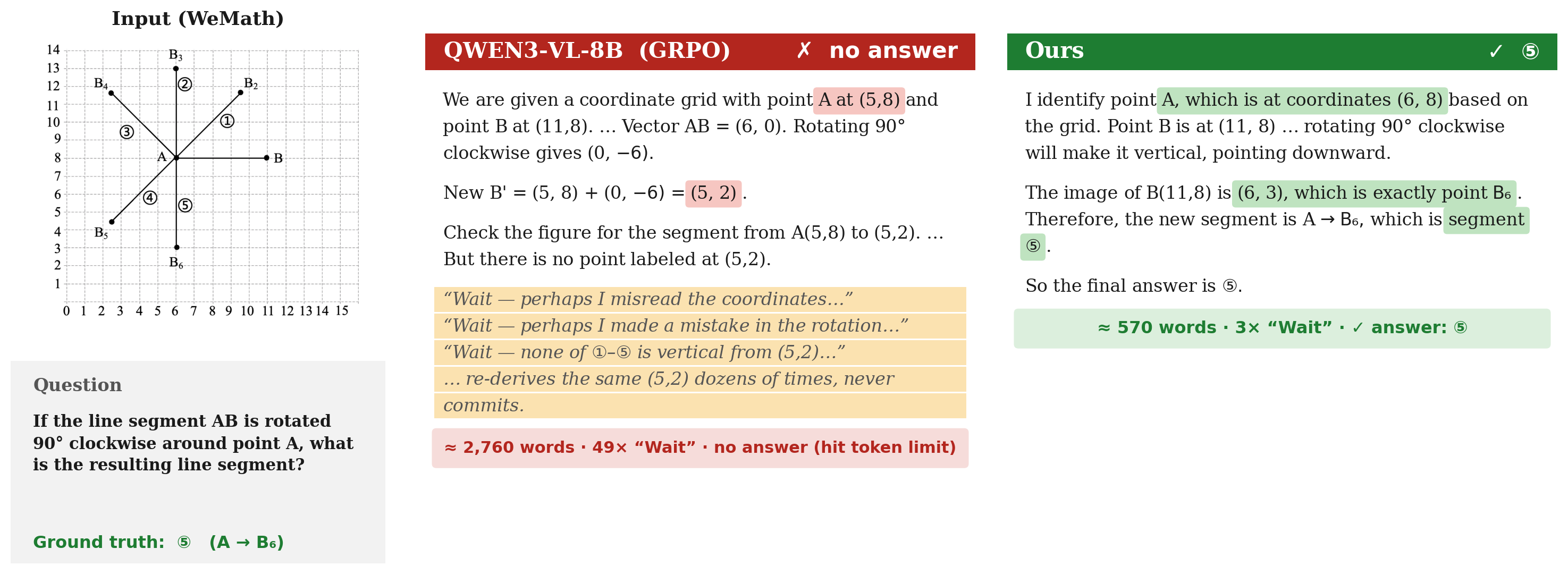}
\caption{\textbf{Qualitative example (WeMath).} Rotating $AB$ by $90^\circ$. GRPO
misreads $A$ as $(5,8)$ and hits the token limit without an answer; \mtdname{} reads
$A=(6,8)$ and answers correctly.}
\label{fig:qual2}
\end{figure}

\paragraph{From claim verdicts to token advantages.} To act on the tokens that
made a claim, each claim's character span is mapped to the response-token
positions it covers,
\begin{equation}
  \mathcal{T}_{ij} = \{\, t \;:\; s_t \ge \alpha_{ij} \ \wedge\ e_t \le \beta_{ij} \,\},
\end{equation}
where $[\alpha_{ij}, \beta_{ij})$ is the character span of claim $c_{ij}$ and
$[s_t, e_t)$ the character range of token $t$, both measured in the decoded response;
$\mathcal{T}_{ij}$ collects the tokens lying entirely within the claim. Each claim's advantage is
scattered onto its own positions and \emph{added} to the standard reward advantage:
\begin{equation}
  A_{\text{total}}(t) \;=\; A_{\text{reward}}(t) \;+\; \lambda_{\text{cla}}\, A_{\text{cla}}(t),
  \qquad
  A_{\text{cla}}(t) = \hat{A}_{ij} \ \ \text{for the claim } c_{ij} \text{ with } t \in \mathcal{T}_{ij}.
  \label{eq:cla-total}
\end{equation}
Here $A_{\text{reward}}$ is the group-relative advantage of $r_{\text{overall}}$
(\eqref{eq:grpo-adv}) broadcast over description tokens, and $A_{\text{cla}}(t)=0$ for
tokens outside any \textsc{high}-relevance claim.

Equation \ref{eq:cla-total} changes nothing about the optimizer: \claname{} is an
\emph{additive, already group-normalized advantage channel} that leaves the objective
\eqref{eq:grpo} unchanged with $\hat{A}_i$ replaced by $A_{total}$ , and reduces to vanilla GRPO when
$\lambda_{\text{cla}}=0$. What it changes is the \emph{granularity} and \emph{targeting} of the learning signal: the holistic rewards give one number to the whole rollout, whereas \claname{} delivers a separately baselined number to the tokens that stated each claim.

%=====================================================================
\section{Experiments}
\label{sec:experiments}
%=====================================================================

\subsection{Experimental Setup}

\paragraph{Baselines and implementation details.} We compare \mtdname{} against
(i) the zero-shot base model; (ii) a GRPO baseline trained under the same setting as
ours; (iii) VLM-CapCurr~\citep{wu2026seeing} and (iv) PEARL~\citep{zhang2026perceptual}. We conduct experiments across model sizes
and backbones, using the Qwen3-VL family~\citep{qwen3vl2025} at 2B, 4B and 8B (dense) and 30B-A3B (MoE),
and InternVL3.5-8B~\citep{wang2025internvl3}. In \mtdname{}, a single set of weights
serves as both the describer and the reasoner. The GRPO baseline uses 4096 max output tokens, and each \mtdname{} stage uses 2048. To match the total
number of training steps, we train the GRPO baseline for 4 epochs and each
\mtdname{} stage for 2 epochs. All our models are trained on 2 nodes of  8x NVIDIA B200 with 180GB memory. Additional Training data and hyperparameters are detailed in
Appendix~\ref{app:impl_details}.

\paragraph{Benchmarks and evaluation.} \label{par:benchmark_eval} We evaluate on six
standard benchmarks for visual math and logic reasoning benchmarks:
MathVista~\citep{lu2024mathvista}, MathVision~\citep{wang2024mathvision},
MathVerse~\citep{zhang2024mathverse}, WeMath~\citep{qiao2024wemath},
LogicVista~\citep{xiao2024logicvista} and DynaMath~\citep{zou2024dynamath}. Accuracy is
our primary metric; ablations additionally measure VR and LR capability separately
(\Secref{sec:ablation-capability}). %\mtdname{} applies its two stages only during training. 

\paragraph{Inference:} \mtdname{} uses VR--LR split only during training. At inference, the policy receives the image and question and answers in a single CoT call, with the same prompt and decoding settings as the GRPO baseline (more details in 
Appendix~\ref{app:inference}), so \mtdname{} adds no inference cost. Notably, GRPO is optimized directly in this single-pass format, whereas \mtdname{} never is. The policy carries what the stages train, accurate perception and evidence-grounded reasoning, into a format it was never optimized in. Following prior works~\cite{wu2026seeing,zhang2026perceptual} we perform greedy decoding across all our experiments.

\subsection{Results}
\subsubsection{Main Results}
\label{sec:main}
Table~\ref{tab:main} reports accuracy on six benchmarks for five backbones spanning
four scales (2B--30B), dense and Mixture-of-Experts architectures, and two model
families. \mtdname{} improves over the GRPO baseline on every backbone, by $+1.5$ to $+6.1$ average points. On the dense Qwen3-VL models, \mtdname{} improves over GRPO by
$\mathbf{+6.1}$ (2B), $\mathbf{+3.7}$ (4B) and $\mathbf{+2.7}$ (8B) points, winning
on all six benchmarks at every size. On Qwen3-VL-2B, it gains $+10.0$ points over the
base model, against GRPO's $+3.9$. The margin over GRPO narrows as the backbone grows, consistent with stronger base models leaving less headroom on these
benchmarks under the same training data and budget; though it remains positive at every
scale. 

On the Mixture-of-Experts Qwen3-VL-30B-A3B, \mtdname{}
improves over GRPO by $\mathbf{+1.4}$ points, with the largest gains on WeMath,
LogicVista and DynaMath ($+3.2$ to $+4.2$), and trails GRPO only on MathVista and
MathVision. On InternVL3.5-8B~\citep{wang2025internvl3},
\mtdname{} improves over GRPO by $\mathbf{+2.1}$ points and on all six benchmarks,
showing that the benefits are not specific to one family. Notably, these improvements are obtained without the two-stage interface used during training. For completeness, we also report two-stage inference results in appendix \ref{tab:self-loop}.

On Qwen3-VL-8B, \mtdname{}
reaches $75.3$ average accuracy, outperforming PEARL~\citep{zhang2026perceptual},
which adds a perception-checklist reward to an entangled CoT, by $\mathbf{+4.6}$
points, and VLM-CapCurr~\citep{wu2026seeing}, which stages post-training on curated
perception data, by $\mathbf{+7.1}$. \footnote{Both are evaluated on their released Qwen3-VL-8B checkpoints.}

\subsubsection{Qualitative results}
\label{sec:qualitative}
\Figref{fig:qual2} compares GRPO and \mtdname{} on a WeMath example (more is in
Appendix~\ref{app:qual_exam}). The baseline's error is perceptual, not logical: after
one misread coordinate, every later step is consistent but wrong, so no rotation
matches an option. Instead of re-reading the figure, it re-derives the same point
($49$ ``Wait''s) until it runs out of tokens, whereas \mtdname{} needs only three.
This matches \Secref{sec:ablation-capability}: answer-only training does not correct
misread visual facts, and when the evidence is wrong, more reasoning becomes hedging
rather than progress.

\subsection{Ablation studies}
\label{sec:ablation}
\begin{figure}[t]
\centering
\begin{minipage}[t]{0.48\textwidth}
  \centering
  \includegraphics[width=0.9\linewidth]{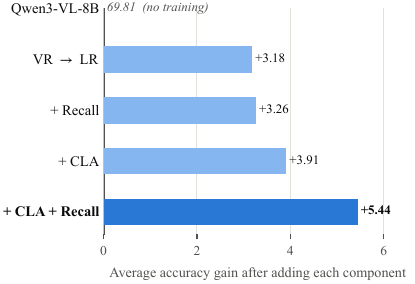}
  \caption{\textbf{Contribution of each component}. Per-benchmark results are in Appendix
  Table~\ref{tab:ablation-components}.}
  \label{fig:ablation-components}
\end{minipage}\hfill
\begin{minipage}[t]{0.48\textwidth}
  \centering
  \includegraphics[width=0.8\linewidth]{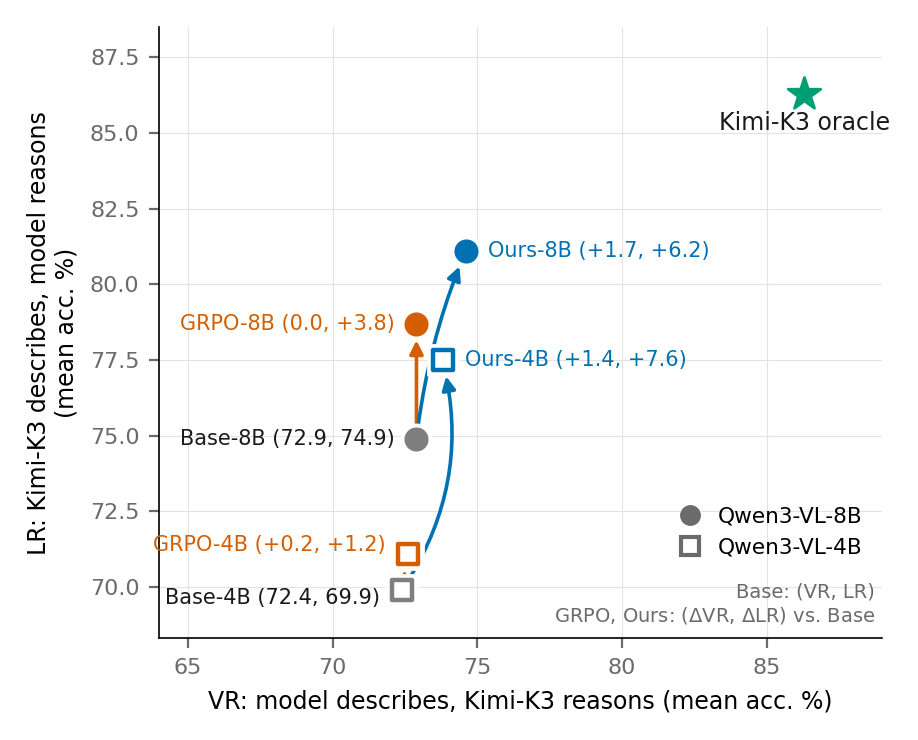}
  \caption{\textbf{Diagnostic probe:} each capability in isolation with an oracle.}
  \label{fig:ablation-capability}
\end{minipage}
\end{figure}

\subsubsection{Contribution of each component}
\label{sec:ablation-components}
Figure~\ref{fig:ablation-components}, isolates the contribution of each component
of \mtdname{} on Qwen3-VL-8B. We start from staged training with only the
bottleneck answer reward in the VR stage (VR$\rightarrow$LR) and add the recall
reward, \claname{}, and both. The LR stage is identical across all variants.

\textbf{Staging alone already helps.} VR$\rightarrow$LR uses no supervision beyond answer correctness, exactly like GRPO, yet improves over the base model by $+3.2$ points and slightly exceeds entangled-CoT GRPO ($72.6$). The benefit is clearer when each capability is measured on its own: staging alone already improves the model both as a describer and as a reasoner (\Secref{sec:ablation-capability}; Appendix Table~\ref{tab:ablation-capability}).

\textbf{Recall and \claname{} are complementary.}  Added separately, recall and \claname{} raise the average by only $+0.1$ and $+0.7$ over VR$\rightarrow$LR. Combined, they give the best average, $75.3$: a $+2.3$ gain over VR$\rightarrow$LR, larger than the sum of the two ($+0.8$). Completeness and correctness are therefore distinct requirements on a description, and accurate VR needs both; the superadditive gain indicates the two signals reinforce rather than duplicate each other.

\subsubsection{Improvement of each capability}
\label{sec:ablation-capability}
We introduce a diagnostic probe that measures each
capability in isolation (\Figref{fig:ablation-capability}). For this analysis only,
we split generation into the two roles and give the other role to Kimi-K3~\citep{team2026kimi}
as a fixed oracle. To measure VR, the model describes and Kimi-K3 reasons; to measure
LR, Kimi-K3 describes and the model reasons from from the Kimi description alone.

\textbf{GRPO improves only LR.} On Qwen3-VL-8B, the descriptions of the GRPO model have the base accuracy ($72.9$), while its reasoning improves by $+3.8$. Qwen3-VL-4B shows the same pattern ($+0.2$ VR, $+1.2$ LR) indicating Answer-only RL using entangled CoT  leaves perception unchanged.

\textbf{\mtdname{} improves both.} As a describer, our 8B model raises Kimi-K3's accuracy by $+1.7$ over both the base and GRPO models. As a reasoner, it gains $+6.2$ over the base, $+2.4$ more than GRPO. The per-benchmark results (Appendix Table~\ref{tab:ablation-capability}) show that most of this LR gain is already obtained by staging alone (VR$\rightarrow$LR). Although LR uses standard GRPO, the model never sees the image and must reason from a visual description only, so every step has to be grounded in stated evidence rather than in shortcuts from the image. Recall and \claname{} increase the LR gain further, even though both act only on the VR stage, because the LR stage is subsequently trained on more complete and correct on-policy descriptions. Better perception and better-grounded reasoning thus reinforce each other as seen in gains of \Secref{sec:ablation-components}. At 4B, \mtdname{} gains $+1.4$ in VR and $+7.6$ in LR. 
\subsubsection{Order of staged training}
\label{sec:ablation-order}

\begin{wraptable}[11]{r}{0.5\linewidth}
\centering
\begin{tabular}{@{}ccc c@{}}
\toprule
Stage 1 & Stage 2 & Mean acc. & Gain \\
\midrule
\multicolumn{2}{@{}c}{Qwen3-VL-8B} & 69.8 & -- \\
\midrule
\multicolumn{2}{@{}c}{Qwen3-VL-8B (GRPO)} & 72.6 & +2.8 \\
LR & VR & 72.1 & +2.3 \\
\textbf{VR} & \textbf{LR} & \textbf{75.3} & \textbf{+5.5} \\
\bottomrule
\end{tabular}
\caption{Order of staged training. Gains are relative to the base model.}
\label{fig:ablation-order}
\end{wraptable}

Our recipe trains VR first and LR second. Table ~\ref{fig:ablation-order}
reverses this order while keeping the rewards and training budget unchanged.
The reversed order (LR$\rightarrow$VR) reaches $72.1$, only $+2.3$ over the
untrained model and $0.5$ below answer-only GRPO, whereas VR$\rightarrow$LR
reaches $75.3$ ($+5.5$). Training the reasoner first therefore gives up most of
the benefit of staging. The drop concentrates on the benchmarks that depend most
on perception, MathVision ($-8.2$) and WeMath ($-5.1$), and the reversed order
is ahead only on MathVista, by $0.2$. A reasoner trained on the descriptions of
an untrained describer learns to compensate for incomplete evidence, and the
later VR stage cannot undo that. Perception should therefore be established
before reasoning is refined.

% \input{5_Ablations}

%=====================================================================

\section{Conclusion} %and Future Work}
%=====================================================================
We introduced \mtdname{}, a staged post-training approach that optimizes VR and LR in stages within a single policy. \mtdname{} uses the description as an information bottleneck, and harnesses final-answer correctness as a supervisory signal for perception, requiring no caption-level labels. The separation further lets each capability be trained with its own reward, and \claname{} turns per-claim verification into a group-relative advantage per visual type that plugs into standard GRPO. On six visual math and logic benchmarks, \mtdname{} improves over
GRPO across model sizes and backbones with average accuracy gains of 1.4–6.1 \%. Our ablations show that staging alone already beats entangled-CoT GRPO, that completeness and correctness
are distinct requirements whose signals (the recall reward and \claname{}) are complementary, and that order matters -- training LR before VR forfeits the benefit of staging, so perception should be established before reasoning is refined. Limitations and future work are discussed in Appendix \ref{app:limitations}.

\section{AI use statement}
Generative AI tools were employed in the course of this work to support multiple activities including writing code for our methods, editing the manuscript's prose and finding existing related works. We additionally relied on such tools to generate and adjust scientific figures. All AI-assisted output was examined by the authors, and we independently confirmed the correctness of the implementation. These tools were not used for intellectual discussion, scientific ideation, or the development of the core research ideas; the conceptual contributions and methodological ideas presented in this work were developed by the authors.

\section{Reproducibility statement}
We provide detailed descriptions of the proposed training procedure, reward design, and optimization objectives in the main paper. The appendix provides additional implementation details, including training configurations, hyperparameters, data and benchmark descriptions, evaluation protocols, detailed prompts and further experimental results. We use publicly available model backbones and evaluation benchmarks and report the settings necessary to reproduce our experiments. 

\section{Ethics statement}
This work develops methods for improving visual and language reasoning in vision--language models. Our experiments use publicly available model backbones and established research benchmarks and do not involve human subjects or the collection of sensitive or personally identifiable information. As with improvements to general-purpose vision--language reasoning, the proposed methods could be incorporated into systems deployed in applications beyond those studied here, including settings in which incorrect model outputs may have real-world consequences. Our experiments are limited to controlled benchmark evaluations and do not establish suitability for such high-stakes deployment.

\bibliography{iclr2027_conference}
\bibliographystyle{iclr2027_conference}

\newpage
\section{Appendix}
\subsection{Objective Function}

\begin{equation}
  J(\theta) = \mathbb{E}\Big[\tfrac{1}{G}\textstyle\sum_{i=1}^{G}
    \tfrac{1}{|o_i|}\textstyle\sum_{t \in o_i}
    \min\big(\rho_{i,t}\hat{A}_i,\ \operatorname{clip}(\rho_{i,t},1-\varepsilon_{\text{low}},1+\varepsilon_{\text{high}})\hat{A}_i\big)\Big],
  \label{eq:grpo}
\end{equation}

where $\rho_{i,t} = \pi_\theta(o_{i,t})/\pi_{\theta_{\text{old}}}(o_{i,t})$

\subsection{Implementation details} 
\label{app:impl_details}
\paragraph{Training data:} For training, We use the data collected in \cite{wu2026seeing}. The dataset consists of five sources for a total of $19{,}555$ examples (see \ref{tab:data}): Math-PUMA~\citep{zhuang2024mathpuma}, GeoQA170K~\citep{chen2021geoqa}, CLEVR-Math~\citep{lindstrom2022clevrmath,johnson2017clevr}, ArxivQA~\citep{li2024arxivqa}, and DOCCI~\citep{onoe2024docci}. Following \cite{kumar2026dive}, we perform a difficulty signal filtering using $16$ rollouts from the base model and computing $\text{pass\_rate}@16 \in [0,1]$; examples that are already solved ($\text{pass\_rate}@16 = 1$) are filtered out, leaving $13{,}862$ training samples. Table \ref{tab:data} shows training data stats.

\begin{table}[!ht]
\centering
\caption{ Training data mixture. Difficulty filtering ($\text{pass\_rate}@16{=}1$
removed) reduces $16{,}195\!\rightarrow\!13{,}862$ samples.}
\label{tab:data}
\begin{tabular}{@{}lr@{}}
\toprule
Source & \# examples \\
\midrule
Math-PUMA (synthesis)      & 6{,}696 \\
GeoQA170K                  & 6{,}499 \\
CLEVR-Math                 & 2{,}000 \\
ArxivQA ($2\times$ downsampled) & 1{,}000 \\
DOCCI ($2\times$ downsampled)   & 3{,}360 \\
\midrule
Total                      & 19{,}555 \\
After difficulty filtering & 13{,}862 \\
\bottomrule
\end{tabular}
\end{table}

\paragraph{Training details:} We use EasyR1 \citep{zheng2025easyr1} framework for all our training. The vision tower is frozen throughout; only the language model is updated. Images are bounded to $[262{,}144,\,1{,}048{,}576]$ pixels. We use a global batch size of $256$ and $8$ rollouts per sample using temperature of 1.0. With $13{,}862$ training examples this gives $54$ optimizer steps per epoch; each \mtdname{} phase runs $2$ epochs ($108$ steps), and the GRPO baseline runs $4$ epochs ($216$ steps). For the VR stage, we set
$\lambda_{\text{recall}}=0.2$ and $\lambda_{\text{ans}}=0.8$ in
\eqref{eq:combined-reward}, and $\lambda_{\text{cla}}=0.5$ in \eqref{eq:cla-total};
the LR stage follows the standard GRPO setting. We use Claude Sonnet
4.6~\citep{anthropic2026claudesonnet46} for all judge calls. We use an LLM judge for the final-answer correctness check, as symbolic verification is not robust \cite{wang2026moca}. Based on DAPO \citep{yu2025dapo}, we disable KL penalty and do asymmetric clipping with $\varepsilon_{\text{low}}=0.2$,
$\varepsilon_{\text{high}}=0.28$. These hyperparameter are consistent across all our experiments including GRPO baseline. For CLA-GRPO decomposition we use sentence level decomposition with PySBD \citep{sadvilkar-neumann-2020-pysbd}, and each component's character span is mapped to response tokens through the tokenizer's offset mapping, keeping only tokens fully contained in the span. A maximum of 60 claims are used for CLA-GRPO and other receive a $A_{cla} = 0$ to penalize long generations.

\paragraph{Inference setup}
\label{app:inference}
All models are evaluated under the same budget: a maximum of $16{,}384$ new
tokens and greedy decoding, with one generation per question. All inference including \mtdname{} are done using a standard CoT followed by a final answer, with the inference time prompt shown below. Rather than matching answer strings symbolically, we score a
response with an LLM judge that is shown the gold answer and the
model's full generation, and decides whether the generation reaches the correct answer.

\newtcblisting{promptbox}[1]{breakable, colback=white, colframe=black!45,
  title=#1, listing only, listing options={basicstyle=\ttfamily\scriptsize,
  breaklines=true, columns=fullflexible, keepspaces=true, showstringspaces=false}}

\begin{promptbox}{\textbf{Inference prompt}}
<question>

You FIRST think about the reasoning process as an internal monologue and then provide the final answer. The reasoning process MUST BE enclosed within <reason> </reason> tags. The final answer MUST BE put in \boxed{} (for a multiple-choice question, put only the option letter inside \boxed{}).
\end{promptbox}

\subsection{Full ablation results}
\label{app:ablation}
Tables~\ref{tab:ablation-components},~\ref{tab:ablation-capability}
and~\ref{tab:ablation-order} list the per-benchmark accuracies behind
Figures~\ref{fig:ablation-components},~\ref{fig:ablation-capability}
and Table ~\ref{fig:ablation-order}.

\begin{table}[t]
\centering
\small
\caption{\textbf{Order of staged training} on Qwen3-VL-8B, per benchmark; the
summary is Table ~\ref{fig:ablation-order}. Both orders use the same VR-stage
rewards (CLA + recall). $\Delta$ is the change in average accuracy relative to
GRPO. Best per column in \textbf{bold}.}
\label{tab:ablation-order}
\setlength{\tabcolsep}{3.2pt}
\begin{tabular}{lcccccccc}
\toprule
Method & MVista & MVision & MVerse & WeMath & LogicVista & DynaMath & \textbf{AVG} & $\Delta$ \\
\midrule
Qwen3-VL-8B (GRPO)                   & 77.9 & 64.5 & 67.8 & 84.6 & 65.6 & 75.3 & 72.6 & --- \\
\midrule
LR $\rightarrow$ VR                  & \textbf{80.0} & 61.2 & 68.8 & 82.3 & 62.4 & 77.7 & 72.1 & \dneg{$-0.5$} \\
\rowcolor{lightgray}
VR $\rightarrow$ LR (\mtdname{})     & 79.8 & \textbf{69.4} & \textbf{69.8} & \textbf{87.4} & \textbf{66.0} & \textbf{79.1} & \textbf{75.3} & \dpos{$\mathbf{+2.7}$} \\
\bottomrule
\end{tabular}
\end{table}

\subsection{Worked example of the recall reward}
\label{app:recall-example}

\begin{figure}[h]
\centering
\includegraphics[width=\textwidth]{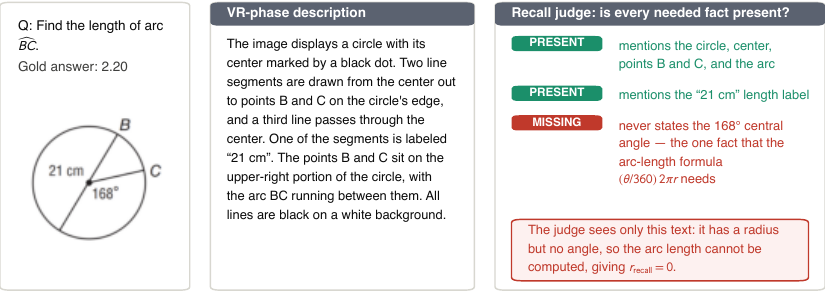}
\caption{\textbf{The recall reward on a worked example.} \emph{Left:} the
question and image (arc length of $\widehat{BC}$). \emph{Center:} the description
produced by the VR phase. \emph{Right:} a text-only judge, which never sees the
image, checks whether the description alone contains every fact the question
needs. It finds the circle, center, points and the ``21\,cm'' label, but the
description \emph{never states the $168^\circ$ central angle}---the one fact the
arc-length formula requires---so the description is insufficient and
$r_{\text{recall}}=0$.}
\label{fig:vrreward}
\end{figure}

\begin{table}[t]
\centering
\small
\caption{\textbf{Contribution of each component} on Qwen3-VL-8B. All staged
variants share the same LR stage and differ only in the rewards used in the VR
stage. $\Delta$ is the change in average accuracy relative to the untrained
Qwen3-VL-8B. The best result per column is shown in \textbf{bold}.}
\label{tab:ablation-components}
\setlength{\tabcolsep}{3.2pt}
\begin{tabular}{lcccccccc}
\toprule
Method & MVista & MVision & MVerse & WeMath & LogicVista & DynaMath & \textbf{AVG} & $\Delta$ \\
\midrule
Qwen3-VL-8B                          & 77.2 & 63.2 & 63.8 & 79.9 & 60.4 & 74.4 & 69.8 & --- \\
\midrule
VR $\rightarrow$ LR                  & 79.3 & 65.5 & 68.2 & 84.8 & 63.1 & 77.1 & 73.0 & \dpos{$+3.2$} \\
VR + Recall $\rightarrow$ LR         & 79.6 & 64.1 & 67.4 & 84.9 & 65.1 & 77.3 & 73.1 & \dpos{$+3.3$} \\
VR + CLA $\rightarrow$ LR            & \textbf{80.1} & 61.2 & \textbf{70.9} & \textbf{87.9} & 65.3 & 76.9 & 73.7 & \dpos{$+3.9$} \\
\rowcolor{lightgray}
VR + CLA + Recall $\rightarrow$ LR   & 79.8 & \textbf{69.4} & 69.8 & 87.4 & \textbf{66.0} & \textbf{79.1} & \textbf{75.3} & \dpos{$\mathbf{+5.4}$} \\
\bottomrule
\end{tabular}
\end{table}

\begin{table}[t]
\centering
\small
\caption{\textbf{Improvement of each capability} on Qwen3-VL-8B. \textbf{(a)} VR:
the describer is varied and the reasoner is fixed to Kimi-K3. \textbf{(b)} LR: the
describer is fixed to Kimi-K3 and the reasoner is varied. The oracle row uses
Kimi-K3 for both roles and is shared by both blocks. $\Delta$ is the change in
mean accuracy relative to the base model. Within each block, best per column (excluding the
oracle) in \textbf{bold}, second best \underline{underlined}.}
\label{tab:ablation-capability}
\setlength{\tabcolsep}{3.2pt}
\begin{tabular}{lcccccccc}
\toprule
Method & MVista & MVision & MVerse & WeMath & LogicVista & DynaMath & \textbf{Mean} & $\Delta$ \\
\midrule
\multicolumn{9}{l}{\emph{(a) Varying the describer (reasoner: Kimi-K3)}}\\
\textcolor{gray}{Kimi-K3 (oracle)}   & \textcolor{gray}{86.4} & \textcolor{gray}{84.9} & \textcolor{gray}{85.9} & \textcolor{gray}{89.8} & \textcolor{gray}{80.1} & \textcolor{gray}{90.6} & \textcolor{gray}{86.3} & \\
Qwen3-VL-8B                          & 75.6 & 66.1 & 72.8 & 80.5 & \textbf{66.2} & 75.9 & 72.9 & $\phantom{+}0.0$ \\
Qwen3-VL-8B (GRPO)                   & 76.3 & \underline{66.8} & \textbf{74.6} & 80.3 & 63.5 & 76.0 & 72.9 & $\phantom{+}0.0$ \\
VR $\rightarrow$ LR                  & \underline{77.0} & 64.8 & 73.7 & 81.7 & \underline{65.5} & \textbf{78.3} & \underline{73.5} & \dpos{$\underline{+0.6}$} \\
VR + Recall $\rightarrow$ LR         & 76.0 & 64.7 & 74.0 & \underline{81.8} & 64.4 & 75.6 & 72.7 & \dneg{$-0.2$} \\
\rowcolor{lightgray}
VR + CLA + Recall $\rightarrow$ LR   & \textbf{79.6} & \textbf{67.4} & \underline{74.4} & \textbf{83.1} & \textbf{66.2} & \underline{76.8} & \textbf{74.6} & \dpos{$\mathbf{+1.7}$} \\
\midrule
\multicolumn{9}{l}{\emph{(b) Varying the reasoner (describer: Kimi-K3)}}\\
\textcolor{gray}{Kimi-K3 (oracle)}   & \textcolor{gray}{86.4} & \textcolor{gray}{84.9} & \textcolor{gray}{85.9} & \textcolor{gray}{89.8} & \textcolor{gray}{80.1} & \textcolor{gray}{90.6} & \textcolor{gray}{86.3} & \\
Qwen3-VL-8B                          & 80.1 & 63.2 & 69.5 & 83.6 & 65.8 & 87.5 & 74.9 & $\phantom{+}0.0$ \\
Qwen3-VL-8B (GRPO)                   & 83.0 & 68.1 & 74.0 & \underline{90.6} & 69.1 & 87.2 & 78.7 & \dpos{$+3.8$} \\
VR $\rightarrow$ LR                  & \underline{84.2} & \underline{69.7} & \textbf{76.6} & \textbf{91.0} & \underline{69.8} & 88.6 & \underline{80.0} & \dpos{$\underline{+5.1}$} \\
VR + Recall $\rightarrow$ LR         & 82.7 & 69.1 & 74.6 & 89.7 & \textbf{71.6} & \underline{88.8} & 79.4 & \dpos{$+4.5$} \\
\rowcolor{lightgray}
VR + CLA + Recall $\rightarrow$ LR   & \textbf{85.5} & \textbf{72.7} & \underline{76.3} & \textbf{91.0} & \textbf{71.6} & \textbf{89.3} & \textbf{81.1} & \dpos{$\mathbf{+6.2}$} \\
\bottomrule
\end{tabular}
\end{table}

\subsection{Two-pass evaluation}
\label{app:self-loop}
All results in the main paper use single-pass CoT inference. Table~\ref{tab:self-loop}
instead evaluates each model in the two-pass format used during training: the model
first describes the image and then answers from its own description, without the
image. \mtdname{} improves over the base model by $+8.1$ points on Qwen3-VL-8B and
$+10.5$ on Qwen3-VL-4B, $3.9$ and $8.2$ points more than GRPO. The two-pass format
is thus where the gains of VR and LR add up most visibly: the model must reason from
its own description, so the modest VR gain measured with Kimi-K3 as the reasoner
(\Secref{sec:ablation-capability}) matters.

Two-pass accuracy is lower than single-pass accuracy for every model, including the
base model ($64.0$ vs.\ $69.8$ on Qwen3-VL-8B). This drop is a property of the format
rather than of training: the description is an information bottleneck, so any visual
fact it omits or misstates cannot be recovered, and the reasoner cannot look at the
image again. \mtdname{} loses the least ($-3.2$ points on Qwen3-VL-8B, vs.\ $-4.4$ for
GRPO and $-5.8$ for the base model; $-4.1$ vs.\ $-8.6$ and $-9.8$ on Qwen3-VL-4B),
consistent with its descriptions being more self-sufficient. \mtdname{} leads in both
formats: in the two-pass format it is trained in, and in the single-pass format that
GRPO is trained in. Since single-pass inference is both cheaper and more accurate, we
use it for all main results.

\subsection{Phase prompts}
\label{app:prompts}
We use one fixed prompt per phase for all training and diagonistic evaluation. The VR-phase
prompt asks the model to state only the visual content relevant to the question,
without reasoning towards an answer; the LR-phase prompt asks the model to answer
from the question and the description alone, without access to the image.

\begin{promptbox}{\textbf{VR phase (describer) prompt}}
You are a careful visual perceiver. You will be shown an image and a
question. Your ONLY job is to produce a faithful, detailed visual
description of the image that is relevant to the question. You MUST
NOT attempt to solve, infer the answer, or perform reasoning beyond
direct observation.

# What to describe

Describe everything in the image that is relevant to the question.
Cover, when present:
- Objects, shapes, and geometric primitives (points, lines, segments,
  rays, angles, arcs, circles, polygons, curves, axes, gridlines).
- Text content exactly as it literally appears in the image
  (character-for-character, including capitalization and punctuation).
- Every numerical value and unit visible in the image, and where each
  value is placed. Aim to preserve ALL numeric tokens that appear on
  the figure, especially side lengths, angle measures, coordinates,
  tick values, and axis labels -- do not paraphrase them away.
- Variables and formulas as they are rendered on the figure. Wrap math
  expressions and single-letter variables in inline LaTeX using
  `$...$` (e.g. `$\angle ABC$`, `$x = 3$`, `$\frac{a}{b}$`,
  `$\overline{PQ}$`, `$60^{\circ}$`).
- Spatial relationships and layout using explicit positional language:
  `upper-left`, `upper-right`, `lower-left`, `lower-right`, `top`,
  `bottom`, `center`, `left of`, `right of`, `above`, `below`,
  `between`, `inside`, `outside`, `adjacent to`, `aligned with`.
  Anchor each labeled element to a position or to another element.
- Fine visual cues: `tick marks`, `hash marks`, `arrowheads`, `dashed
  line`, `dotted line`, `solid line`, `bold` / `italic` text, colors
  (only when visually distinctive: `red`, `blue`, `green`, `pink`,
  `orange`, `gray`, shading, fill patterns), congruence marks,
  right-angle squares, parallel-line ticks, decorative styling.
- Counts and ordering (e.g. "three points labeled A, B, C from left
  to right"; "four bars, tallest on the right").
- For charts/plots: axis titles, axis ranges, tick values, legend
  entries, series names, and the qualitative shape of each series
  (increasing, decreasing, peaked, flat) -- but do not compute values.
- For tables: headers, row labels, and cell entries verbatim.

# Style and calibration

- Use bulleted structure (lines beginning with `- `) for enumerable
  visual facts (labels, coordinates, measurements, series). Use short
  prose paragraphs only for overall layout summary.
- Be epistemically calibrated: use hedges such as `appears to be`,
  `looks like`, `seems`, `roughly`, `approximately` when a value,
  color, or relationship is estimated from pixels rather than read
  from an explicit label. Do NOT hedge on text/numbers that are
  printed and legible -- quote those exactly.
- Describe what is actually drawn. Do NOT invent stock phrases like
  "coordinate plane", "clockwise order", "counter-clockwise order",
  "not labeled", "no measurement", "no markings" unless that absence
  is genuinely relevant to answering the question and is a real
  feature of the image.
- If something is illegible or ambiguous, say so explicitly and, if
  possible, describe what you can see (e.g. "a two-digit number,
  first digit appears to be `4`, second is unclear").

# Output contract

- Output a SINGLE block of plain text (bullets allowed inside it).
- Target length: roughly 800-1500 characters. Shorter is fine only if
  the image is trivially simple; longer is fine only if the image is
  unusually dense. Never exceed 2500 characters. Never repeat the
  same phrase or bullet -- if you find yourself repeating, STOP.
- Do NOT output a `Final Answer`, `Answer:`, `Therefore`, a solution,
  a computed result, or any conclusion. Do NOT solve the problem.
- Do NOT introduce facts, measurements, or labels not visible in the
  image. Do NOT infer values that are not explicitly shown.

## Question:
<question>

## Visual Description:
\end{promptbox}

\begin{promptbox}{\textbf{LR phase (reasoner) prompt}}
You are given (a) a question, (b) a visual description of the image
that another model produced. Using ONLY the visual
description and the question, solve the problem
step by step.

Follow these rules strictly:
- Commit to ONE interpretation of the figure on your first read and
  solve under that interpretation. Do not enumerate alternative
  layouts or restart from a different reading. Do not write
  "wait", "actually", "hold on", "let me reconsider", or similar
  self-correction openers -- if you notice a mistake, fix that
  specific step and move on.
- Every quantity named in the description (a length, an angle, a
  label like "13" or "6x+7") is real and given for a reason. Use it
  as data. Do not doubt whether it is a length, an angle, or valid.
- Assume the problem is solvable. Only after you have made a
  serious attempt and can identify a specific missing fact should
  you consider refusing. Prefer a best-effort numeric answer under
  the most natural interpretation over refusing.
- Keep the reasoning short and linear: one plan, one derivation,
  one final answer.

If, after a genuine attempt, a specific fact truly is missing and
no natural interpretation lets you proceed, briefly state what is
missing and answer can't be answered.

## Question:
<question>

## Visual Description (from describer):
<description>
\end{promptbox}

\subsection{Judge prompts}
\label{app:judge-prompts}
We use two judges in the VR stage, both instantiated with Claude Sonnet
4.6~\citep{anthropic2026claudesonnet46}, the same model used for answer
evaluation (Appendix~\ref{app:inference}). The \emph{recall} judge is text-only: it
receives the question and the VR-phase description, but never the image, and
decides whether the description alone contains every visual fact needed to answer
the question. The \claname{} judge does see the image, and labels each atomic
claim with a correctness verdict, a relevance verdict and a visual type.

\begin{promptbox}{\textbf{Recall judge prompt (VR phase)}}
You are auditing a visual description for information sufficiency AND for staying within its role. You are NOT given the image. You are given (a) a question that was asked about an image and (b) a description of that image produced by another model. The description's ONLY job is to state the visual facts of the image; a separate reasoning stage (which you do NOT see) is responsible for using those facts to reason about and answer the question.

Decide the score by applying BOTH rules below. The score is 0 if EITHER rule fails.

Rule A -- sufficiency:
- The description must state (explicitly or unambiguously) every visual fact a competent reasoner would need to answer the question. If at least one required piece of visual information is missing, ambiguous, or only partially specified, the description is insufficient.

Rule B -- role (no reasoning / no answer):
- The description must ONLY describe visual elements. It must NOT do the reasoning the later stage is responsible for, and it must NOT state or imply the answer to the question. This includes: giving the final answer, computing or deriving a result, drawing a conclusion, comparing/selecting among options to decide which is correct, or narrating step-by-step reasoning toward the answer. Reporting a visual fact that is directly readable from the image (e.g. a printed label, a number written in the figure, a color, a count of visible objects) is fine and expected; performing the calculation or inference that answers the question is NOT.

Scoring:
- score = 1 -- the description satisfies Rule A (sufficient) AND Rule B (purely descriptive, no reasoning and no answer).
- score = 0 -- the description fails Rule A (insufficient) OR fails Rule B (it contains the answer or performs the reasoning). If the description contains or derives the answer, score 0 even if it is otherwise sufficient.

Important:
- Do NOT judge whether the description is correct -- only sufficiency (Rule A) and role (Rule B). Even a wrong description can be sufficient.
- Do NOT judge whether the question can be answered without an image (text-only) -- assume the question genuinely depends on visual content.
- Numerical values that are literally printed/visible in the image are visual facts and are allowed; a calculation, comparison, or inference performed on those values to reach the answer is reasoning and is NOT allowed.

Question:
<question>

Description:
<description>

Respond with ONLY a single JSON object of this shape:

{"score": 0, "rationale": "<=1 short sentence"}

No markdown, no commentary outside the JSON object, no code fences.
\end{promptbox}

\begin{promptbox}{\textbf{\claname{} judge prompt (claim verification)}}
You are verifying the factual accuracy of a visual description against the image it describes. You ARE given the image. You are also given (a) a question that was asked about the image and (b) a list of components -- individual claims extracted from a model's description of the image. Judge EACH component independently.

For each component, decide:
- correct: 1 if the component is a faithful, accurate statement about the image (it does not contradict what is visible and is supported by the image); 0 if it is false, hallucinated, or contradicts the image.
- relevance: HIGH if the component states visual information that matters for answering the question; LOW if it is accurate/inaccurate but tangential to the question.
- component_type: the single category (from the list below) that best describes what kind of visual claim the component makes.
- rationale: a terse justification of at most 12 words (keep it short).

Allowed component_type values (choose exactly one, verbatim): object_presence, count, color_material, spatial_position, text_label, geometric_relation, line_style, absence, scene_context, other

Category meanings (guidance for the default categories; if a name above is not listed here, judge by its plain meaning):
- object_presence: an object / shape / curve exists, or its identity (e.g. "a circle", "a red cylinder", "a lizard").
- count: a number or quantity of items (e.g. "three points", "ten circles", "there are two segments").
- color_material: color, shade, material, or finish (e.g. red, gray, matte, metallic, reflective).
- spatial_position: location, orientation, or relative position (e.g. left, above, center, quadrant, aligned).
- text_label: labels, tick values, symbols, captions, or axis names (e.g. "labeled A", tick "-3d", overline/angle notation on segments).
- geometric_relation: a quantitative or geometric relation (e.g. parallel, perpendicular, angle, equal, congruent, bisects, coordinate, slope, connects, intersects, forms a triangle).
- line_style: stroke or fill attributes (e.g. solid, dashed, thick, filled, arrowhead, hatching).
- absence: a negated or absent feature (e.g. "no numerical values", "no gridlines", "no colors").
- scene_context: background, surface, lighting, shadow, or gridlines.
- other: anything that fits none of the above.

Rules:
- Judge correctness ONLY against the image, not against the question's answer. A component can be correct (faithful to the image) even if it is irrelevant to the question.
- Judge each component on its own; do not let one component's error affect another's score.
- If a component is vague or unverifiable from the image, mark correct = 0.
- Pick the single best-fitting component_type. If a component makes several claims, pick the type of its PRIMARY claim. If none fit well, use "other".

Question:
<question>

Components (index in brackets):
<indexed list of atomic claims>

Respond with ONLY a single JSON object of this exact shape:

{"components": [{"index": 0, "correct": 0, "relevance": "LOW", "component_type": "other", "rationale": "<=1 short sentence"}]}

Include exactly one entry per input component, using the same index. No markdown, no commentary outside the JSON object, no code fences.
\end{promptbox}

\begin{table}[t]
\centering
\small
\caption{\textbf{Two-phase (self-loop) evaluation.} The model first describes the image
and then answers from its own description, without the image, as in training.
$\Delta$ is the change in mean accuracy relative to the base model of the same size.
Best per column within each backbone in \textbf{bold}.}
\label{tab:self-loop}
\setlength{\tabcolsep}{3.2pt}
\begin{tabular}{lcccccccc}
\toprule
Method & MVista & MVision & MVerse & WeMath & LogicVista & DynaMath & \textbf{Mean} & $\Delta$ \\
\midrule
\multicolumn{9}{l}{\emph{Qwen3-VL-8B}}\\
Base & 69.1 & 54.3 & 58.1 & 73.9 & 57.7 & 71.1 & 64.0 & -- \\
GRPO & 73.4 & 57.2 & 63.7 & 80.3 & 62.4 & 72.2 & 68.2 & \dpos{$+4.2$} \\
\rowcolor{lightgray}
\mtdname{} & \textbf{76.6} & \textbf{62.8} & \textbf{66.4} & \textbf{85.9} & \textbf{65.3} & \textbf{75.9} & \textbf{72.1} & \dpos{$\mathbf{+8.1}$} \\
\midrule
\multicolumn{9}{l}{\emph{Qwen3-VL-4B}}\\
Base & 68.4 & 45.7 & 51.9 & 64.1 & 50.8 & 70.6 & 58.6 & -- \\
GRPO & 69.1 & 49.0 & 58.5 & 68.6 & 50.3 & 69.7 & 60.9 & \dpos{$+2.3$} \\
\rowcolor{lightgray}
\mtdname{} & \textbf{74.3} & \textbf{55.6} & \textbf{64.1} & \textbf{84.3} & \textbf{61.1} & \textbf{75.0} & \textbf{69.1} & \dpos{$\mathbf{+10.5}$} \\
\bottomrule
\end{tabular}
\end{table}

\subsection{Additional Qualitative Examples}
\label{app:qual_exam}
\begin{figure}[t]
\centering
\includegraphics[width=\textwidth]{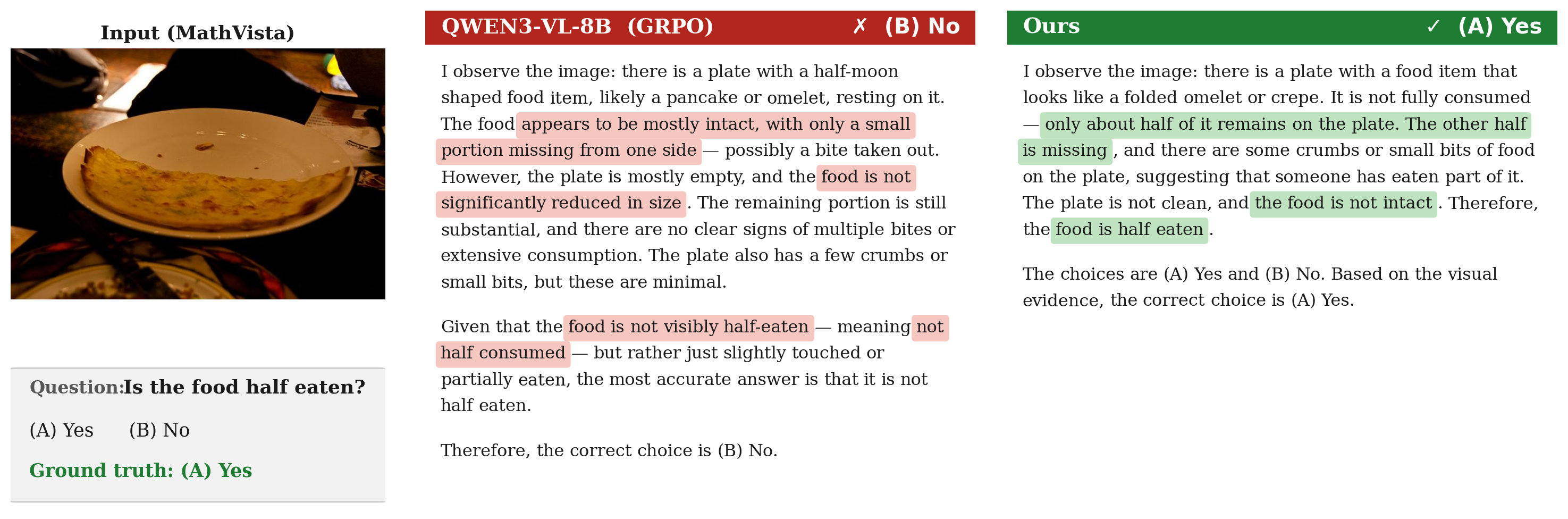}
\caption{\textbf{Qualitative example (MathVista).} \emph{Is the food half
eaten?} The GRPO baseline hedges about the food being ``mostly intact'' and answers
\textbf{(B)~No} (\ding{55}). Our model identifies that only half the food remains
and answers \textbf{(A)~Yes} (\ding{51}), matching the ground truth.}
\label{fig:qual1}
\end{figure}

we show addition quanlitative example in figure \ref{fig:qual1}.

\subsection{Limitations and future work}
\label{app:limitations}

\mtdname{} relies on LLM judges for the recall reward and \claname{}, which adds training cost, and its descriptions still trail those of a frontier model, so perception remains the main source of headroom.  With the vision tower frozen, \mtdname{}'s gains over GRPO shrink as backbones improve, leaving its effectiveness at frontier scale and the need for harder perception-bound data open. Natural next steps are to extend the approach to other domains such as video, and to settings with multiple interleaved rounds of perception and reasoning.

\end{document}